\documentclass[letterpaper]{article}
\usepackage[preprint]{aaai2027}
\usepackage[hyphens]{url}
\usepackage{graphicx}
\usepackage{natbib}
\usepackage{caption}
\usepackage{algorithm}
\usepackage{algorithmic}
\usepackage{booktabs}
\usepackage{multirow}
\usepackage{tabularx}
\usepackage{colortbl}
\usepackage{amsmath}
\usepackage{amssymb}
\usepackage{mathtools}
\usepackage[utf8]{inputenc}

\title{Sample-Adaptive Latent Rewards for Uncertainty-Guided Diffusion Post-Training}
\author{
Rui Li\textsuperscript{\rm 1,\rm 2},
Yuanzhi Liang\textsuperscript{\rm 2},
Ke Hao\textsuperscript{\rm 3,\rm 2},
Ziqiao Weng\textsuperscript{\rm 2},\\
Haibin Huang\textsuperscript{\rm 2},
Chi Zhang\textsuperscript{\rm 2},
XueLong Li\textsuperscript{\rm 2}
}
\affiliations{
\textsuperscript{\rm 1}University of Science and Technology of China, Hefei, China\\
\textsuperscript{\rm 2}Institute of Artificial Intelligence, China Telecom (TeleAI), Shanghai, China\\
\textsuperscript{\rm 3}Shanghai Jiao Tong University, Shanghai, China\\
rui.li@mail.ustc.edu.cn, xuelong\_li@ieee.org
}

\begin{document}

\maketitle

\begin{abstract}
Latent reward models can supervise visual diffusion models without decoding intermediate states into pixel space. This makes alignment with human preferences more efficient. However, existing latent reward models output only scalar scores. They do not estimate the uncertainty of each prediction. The generator therefore cannot determine which feedback is reliable. This can drive optimization in the wrong direction and lead to reward hacking.
We propose \textsc{SURE}, a unified latent-space framework for image and video diffusion models. It learns reward distributions and directly uses their reliability to guide dense post-training. First, we propose sample-adaptive latent reward model (\textsc{SURE-LRM}). It predicts a Gaussian utility for each noisy latent. Its mean predicts the reward score. Its variance reflect the uncertainty of prediction without human annotation. The learned distribution then guides post-training through uncertainty-guided reward feedback learning (\textsc{SURE-REFL}). This method provides uncertainty-guided dense feedback along the denoising trajectory. At selected transitions, \textsc{SURE-REFL} queries the frozen \textsc{SURE-LRM}. It converts detached variance into reliability weights for samples at the same transition. Each weighted reward is backpropagated only through its local transition. The entire process remains in latent space and requires neither pixel-space decoding nor the full denoising graph.
Experiments show that \textsc{SURE-LRM} improves preference prediction over strong baselines. \textsc{SURE-REFL} achieves the sota performance among various metrics and further improves optimization stability. It also achieves the highest VBench quality, semantic, and total scores among the evaluated methods.
\end{abstract}

\section{Introduction}
\label{sec:introduction}

Post-training has become central to aligning visual diffusion models with human preferences. Existing pipelines commonly evaluate preference signals on decoded RGB outputs, introducing repeated VAE decoding into the feedback loop~\cite{prabhudesai2024alignprop,black2024ddpo,liu2025flowgrpo}. Diffusion-native reward models instead score noisy latents directly, avoiding this decoding cost and making intermediate-state supervision possible~\cite{zhang2025lrm,zhang2026drm}. However, existing latent rewards typically output only a scalar score.

DiNa-LRM further advances this direction by explicitly
considering the noise structure of diffusion models\cite{liu2026dina}. Rather than learning rewards only from clean latents,
it trains a timestep-conditioned reward model on states with
different noise levels. It takes the denoising scheduler as uncertainty into preference learning. Higher-noise states are
assigned greater uncertainty as they contain less information. However, noise level alone does not fully determine the
reliability of a reward prediction. DiNa-LRM assigns the
same uncertainty to all latent samples at the same timestep.
In practice, reward difficulty also depends on the specific content being evaluated. At a fixed noise level, one latent
may already reveal a clear subject, while another may remain ambiguous. A relatively noisy but easy sample may therefore support a more reliable prediction than a lower-noise but difficult sample. A second limitation concerns how uncertainty is used. In
DiNa-LRM, noise-dependent uncertainty supports reward
model training, but it is not retained for generator optimization. However, a reward prediction with high uncertainty should have less influence than a reliable prediction. Since unreliable local feedback may repeatedly affect generator updates, even leds to reward hacking.

\begin{figure}[t]
\centering
\includegraphics[width=\columnwidth]{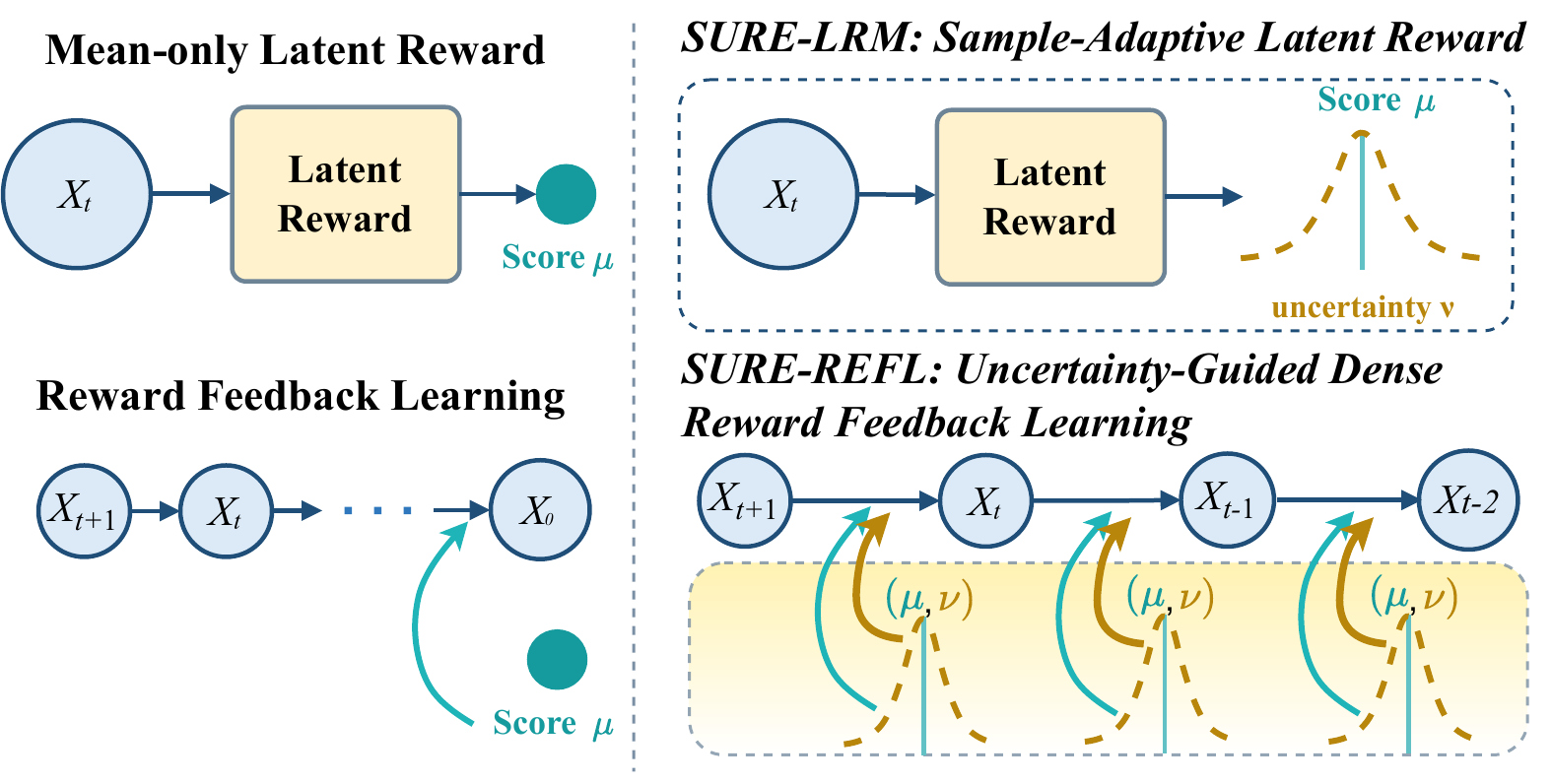}
\caption{
Comparison of reward feedback paradigms.
The mean-only latent rewards cannot express predictive uncertainty. \textsc{SURE-LRM} predicts a latent reward and sample-adaptive uncertainty. REFL fails to leverage predictive uncertainty, while \textsc{SURE-REFL} uses detached uncertainty to control dense local feedback.
}
\label{fig:motivation_overview}
\end{figure}

To address these limitations, we propose \textsc{SURE}, a unified framework for sample-adaptive latent reward uncertainty modeling and diffusion post-training. The framework consists of two parts, \textsc{SURE-LRM} and \textsc{SURE-REFL}. First, \textsc{SURE-LRM} builds upon the diffusion uncertainty prior to predict residual uncertainty conditioned on the latent, prompt, and timestep. It predicts a mean reward and sample-dependent variance using a heteroscedastic preference objective learned from standard pairwise preferences without additional uncertainty or confidence annotations. Variance regularization controls uncertainty inflation, while margin regularization maintains score separation. Additionally, \textsc{SURE-REFL} leverages the frozen \textsc{SURE-LRM} directly in latent space. It queries multiple denoising transitions and backpropagates each reward only through its local transition. Detached uncertainty controls the relative contribution of feedback by comparing samples evaluated at the same transition for both image and video generators. Thus, reward learning and generator alignment remain in latent space without VAE decoding or retention of the full denoising graph.

Experiments indicate \textsc{SURE-LRM} achieves state of art  preference prediction over multi-backbones as latent reward model.  \textsc{SURE-REFL} delivers meaningful performance across diverse evaluation metrics on both image and video post-training. Moreover, \textsc{SURE-REFL} exhibits more stable training behavior and reduced reward hacking. Our contributions are summarized as follows:
\begin{itemize}

\item We formulate sample-adaptive reward uncertainty in latent
space through a preference model. We derive its
variance gradients to show how does it provides an
uncertainty-learning signal.

\item We propose \textsc{SURE-LRM}, which learns sample-adaptive latent reward uncertainty from standard pairwise preferences without additional confidence annotations.

\item We develop \textsc{SURE-REFL}, a unified image--video post-training method that allocates dense latent feedback using detached uncertainty without VAE decoding.

\item Experiments show state-of-the-art post-training performance in the evaluated settings across both image backbones and video backbone.
\end{itemize}

\section{Related Work}
\label{sec:related_work}

\paragraph{Visual reward models.}
Early learned visual rewards map decoded image--text pairs to scalar utilities. ImageReward trains such a scorer from human comparisons~\cite{xu2023imagereward}, while PickScore, HPSv2, and MPS expand user-preference data, cross-distribution evaluation, or multi-dimensional assessment~\cite{pickscore,hpsv2,mps}. More recent VLM-based systems increase model and task coverage: UnifiedReward evaluates both multimodal understanding and generation, UnifiedReward-Think adds explicit reward reasoning, and HPSv3 combines a larger preference corpus with uncertainty-aware ranking~\cite{unifiedreward,unifiedrewardthink,ma2025hpsv3}. These pixel-space evaluators require VAE decoding, which is costly when rewards are queried at many denoising states. Diffusion-native approaches instead evaluate intermediate latents. LRM repurposes diffusion features to score arbitrary noisy states for step-level preference optimization~\cite{zhang2025lrm}; DiNa-LRM learns a general-purpose latent reward with a noise-calibrated Thurstone likelihood and multi-noise inference~\cite{liu2026dina}; and DRM employs a diffusion backbone for stepwise guidance in group relative policy optimization (GRPO) and sampling~\cite{zhang2026drm}. These methods establish that noisy states are scoreable, but they do not expose sample-specific reward reliability to the downstream optimizer.

\paragraph{Diffusion preference optimization.}
Post-training methods differ in how they convert feedback into generator updates. AlignProp and DRaFT propagate gradients from differentiable rewards through full or truncated denoising graphs~\cite{prabhudesai2024alignprop,clark2024draft}. DDPO casts denoising as a multi-step Markov decision process and applies policy-gradient reinforcement learning to terminal rewards~\cite{black2024ddpo}, whereas Diffusion-DPO directly optimizes offline pairwise preferences through a diffusion-specific DPO objective~\cite{wallace2024diffusiondpo}. Building on group relative policy optimization (GRPO)~\cite{shao2024deepseekmath}, Flow-GRPO extends online optimization to flow-matching models~\cite{liu2025flowgrpo}, DanceGRPO applies it across image and video generation tasks~\cite{xue2025dancegrpo}, and ViPO redistributes scalar feedback into spatially and temporally structured advantages~\cite{ni2025vipo}. For video generation, Process Reward Feedback Learning (PRFL) repurposes a pretrained video generator as a noisy-latent process reward model and backpropagates preference feedback through the denoising chain without VAE decoding~\cite{mi2025prfl}. These methods improve the optimization or credit-assignment rule, but dense feedback additionally requires the reliability of intermediate reward predictions to be estimated and used during generator updates.

\paragraph{Uncertainty-aware reward learning.}
Confidence-aware reward optimization shows that visual-model fine-tuning can benefit from attenuating unreliable reward signals~\cite{kim2024confidence}, while probabilistic uncertain reward modeling represents reward as a distribution rather than a point estimate~\cite{sun2025purm}. We bring this principle to diffusion-native visual rewards: \textsc{SURE-LRM} learns sample-dependent relative reliability from ordinary pairwise preferences, and \textsc{SURE-REFL} uses it to weight dense local feedback among states evaluated at the same transition.

\section{Method}
\label{sec:method}

\begin{figure}[t]
    \centering
    \includegraphics[width=\columnwidth]{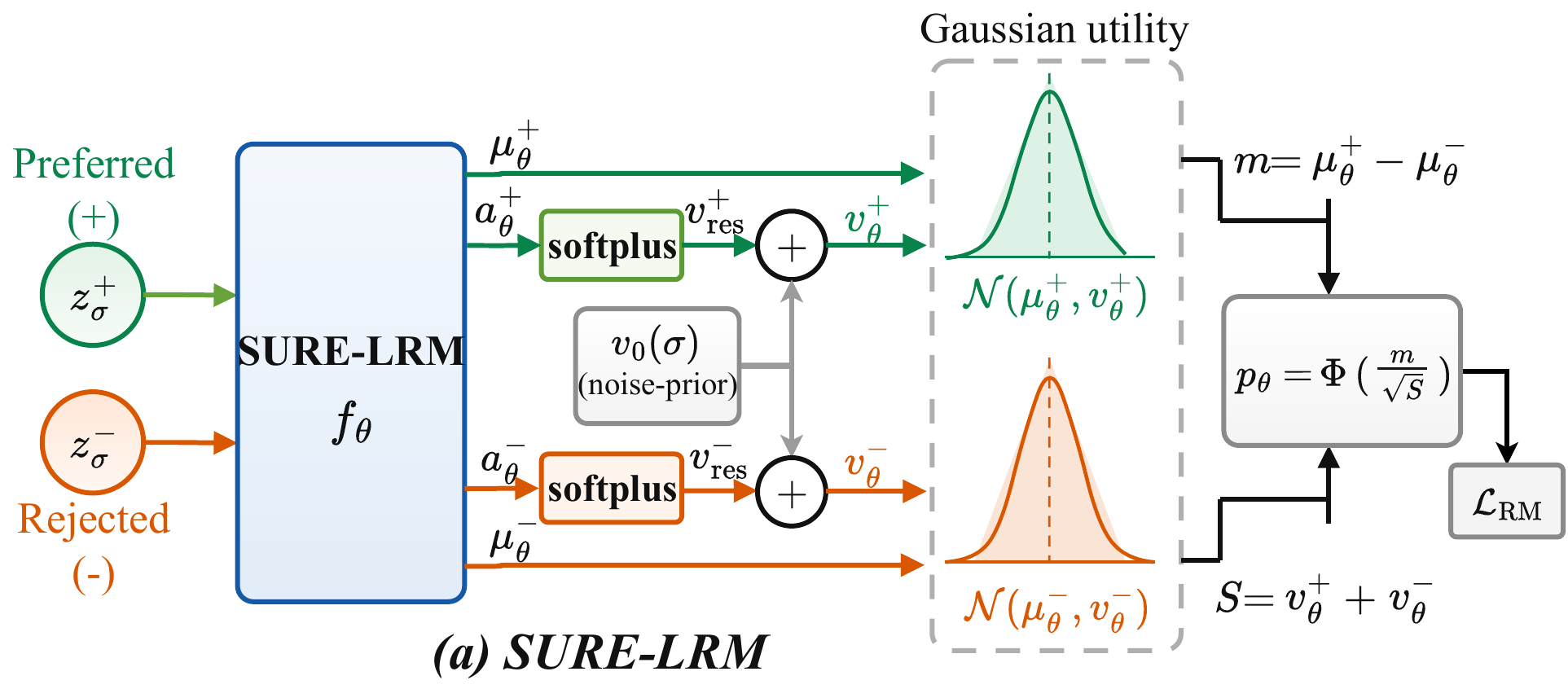}
    \caption{Overview of \textsc{SURE-LRM}. The latent reward backbone predicts a mean reward and sample-adaptive residual uncertainty from the noisy latent, prompt, and timestep. The residual is combined with the noise-dependent prior to form the total predictive variance.}
    \label{fig:sure_lrm_pipeline}
\end{figure}

\begin{figure}[t]
    \centering
    \includegraphics[width=\columnwidth]{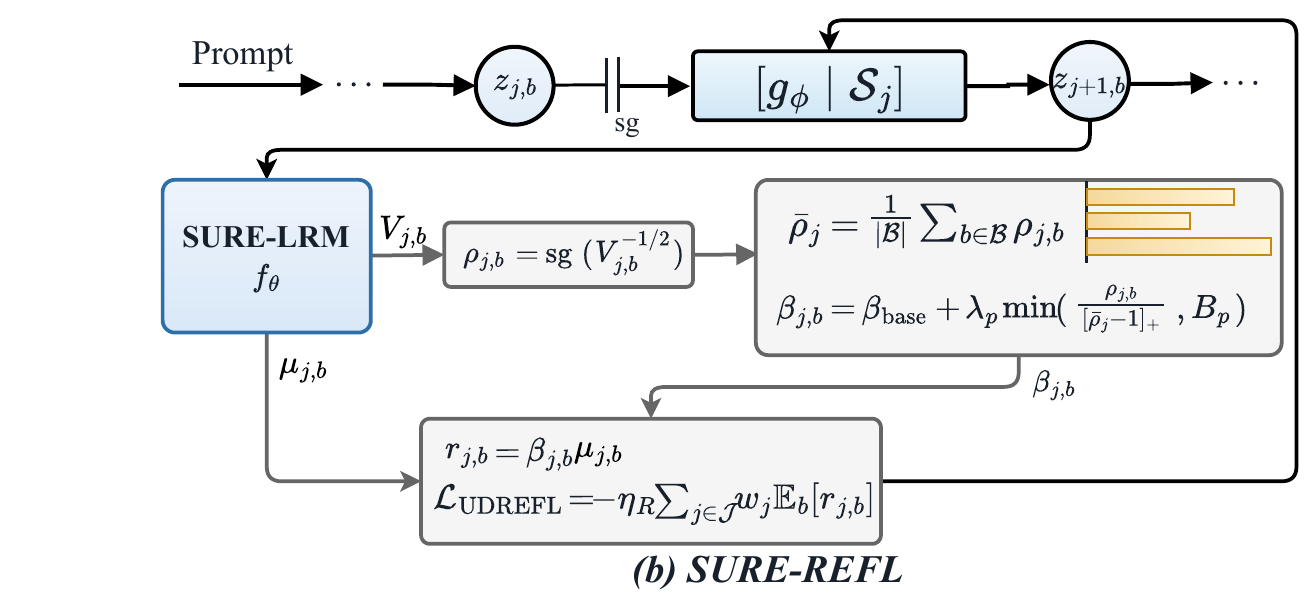}
    \caption{Overview of \textsc{SURE-REFL}. The frozen \textsc{SURE-LRM} evaluates selected denoising transitions and converts detached uncertainty into matched-transition reliability weights. Each reward is backpropagated only through its local transition.}
    \label{fig:sure_refl_pipeline}
\end{figure}

To enable uncertainty-aware reward modeling and efficient reward-guided post-training in latent space, we first develop \textsc{SURE-LRM}, which predicts a preference score together with sample-adaptive uncertainty for each noisy latent. We then freeze the reward model and introduce \textsc{SURE-REFL} to allocate dense reward feedback across denoising transitions using detached uncertainty. Figures~\ref{fig:sure_lrm_pipeline} and~\ref{fig:sure_refl_pipeline} illustrate the two components, respectively. Due to space constraints, the detailed proof is provided in the supplementary material.

\subsection{Preliminaries}
\label{sec:method_preliminaries}

Given a prompt $c$ and a preferred--rejected latent pair $\mathbf z_0^+ \succ \mathbf z_0^-$, we perturb candidate $i\in\{+,-\}$ at noise level $\sigma\in[0,1]$ using rectified-flow interpolation:
\begin{equation}
\mathbf z_\sigma^i=(1-\sigma)\mathbf z_0^i+\sigma\boldsymbol\epsilon^i,
\qquad \boldsymbol\epsilon^i\sim\mathcal N(\mathbf{0},\mathbf I).
\label{eq:noisy_latent}
\end{equation}
The reward model receives $(\mathbf z_\sigma^i,c,t_\sigma)$, where $t_\sigma$ is the scheduler timestep associated with $\sigma$. Under the Thurstone random-utility model~\cite{thurstone1927law}, we write $U_\theta^i=\mu_\theta^i+\eta_\theta^i$, with $\eta_\theta^i\sim\mathcal N(0,v_\eta^i(\sigma))$. Here, $\mu_\theta^i$ is the mean reward and $\eta_\theta^i$ is an unobserved utility residual. Following DiNa-LRM~\cite{liu2026dina}, we retain the noise-dependent variance prior
\begin{equation}
v_0(\sigma)=2\sigma^2+\sigma_0^2,
\label{eq:noise_variance_prior}
\end{equation}
where $\sigma_0^2$ is a small variance floor. This prior captures the average loss of preference information as noise increases, but cannot distinguish candidates at the same noise level. We therefore model the candidate-specific variance not explained by this shared prior.

\subsection{\textsc{SURE-LRM}:Sample-Adaptive Latent Reward}
\label{sec:sample_adaptive_uncertainty}

\paragraph{Sample-adaptive reward distribution.}
For each noisy candidate, \textsc{SURE-LRM} predicts a mean score and an unconstrained residual-variance parameter:
\begin{equation}
f_\theta(\mathbf z_\sigma^i,c,t_\sigma)
=(\mu_\theta^i,a_\theta^i), \qquad i\in\{+,-\}.
\label{eq:sure_reward_outputs}
\end{equation}
Its non-negative, candidate-specific residual variance is
\begin{equation}
v_{\eta,\mathrm{res}}^i
=\operatorname{softplus}(a_\theta^i)+\epsilon_v,
\label{eq:residual_variance}
\end{equation}
and the total utility variance is
\begin{equation}
v_\eta^i(\sigma)=v_0(\sigma)+v_{\eta,\mathrm{res}}^i.
\label{eq:total_variance}
\end{equation}
Here, $\epsilon_v>0$ is a numerical floor. The term is residual because it captures uncertainty beyond $v_0(\sigma)$; it is predicted from the latent, prompt, and timestep without uncertainty annotations.

\paragraph{Heteroscedastic preference learning.}
Following the standard probit formulation for pairwise preference learning~\cite{chu2005preference}, we model the two utility residuals as conditionally independent Gaussians. Their difference has mean $m=\mu_\theta^+-\mu_\theta^-$ and variance
\begin{equation}
\begin{aligned}
S&=v_\eta^+(\sigma)+v_\eta^-(\sigma)\\
 &=2v_0(\sigma)+v_{\eta,\mathrm{res}}^++v_{\eta,\mathrm{res}}^-.
\end{aligned}
\label{eq:pair_variance}
\end{equation}
Letting $\Phi$ denote the standard Gaussian cumulative distribution function, the probability of the annotated preference is
\begin{equation}
p_\theta
=\Pr(U_\theta^+>U_\theta^-)
=\Phi\!\left(\frac{m}{\sqrt S}\right).
\label{eq:sure_preference_probability}
\end{equation}
We optimize its negative log-likelihood:
\begin{equation}
\mathcal L_{\mathrm{pref}}
=-\mathbb E[\log p_\theta].
\label{eq:sure_preference_loss}
\end{equation}
For a single pair, define $q=m/\sqrt S$ and $\ell_{\mathrm{pref}}=-\log\Phi(q)$. Its gradients are
\begin{align}
\frac{\partial\ell_{\mathrm{pref}}}{\partial S}
&=\frac{\varphi(q)}{\Phi(q)}\frac{m}{2S^{3/2}},
\nonumber\\
\frac{\partial\ell_{\mathrm{pref}}}{\partial v_{\eta,\mathrm{res}}^i}
&=\frac{\partial\ell_{\mathrm{pref}}}{\partial S},
\nonumber\\
\frac{\partial\ell_{\mathrm{pref}}}{\partial a_\theta^i}
&=\frac{\partial\ell_{\mathrm{pref}}}{\partial v_{\eta,\mathrm{res}}^i}
\operatorname{sigmoid}(a_\theta^i),
\qquad i\in\{+,-\}.
\label{eq:uncertainty_learning_gradient}
\end{align}
Here, $\varphi$ is the standard Gaussian density and $\varphi(q)/\Phi(q)>0$. A positive $m$ makes gradient descent reduce the residual variance and sharpen a correct ordering; a negative $m$ increases it and softens an overconfident error toward probability $1/2$. Thus, the shared state-conditioned head learns relative uncertainty directly from preference labels; at $m=0$, the likelihood provides no first-order variance signal.

The likelihood alone can nevertheless reduce the penalty for an incorrect ranking by inflating its predicted variance. We therefore add the log-variance regularizer
\begin{equation}
\mathcal L_{\mathrm{var}}
=\tfrac12\mathbb E[\log v_\eta^++\log v_\eta^-],
\label{eq:variance_regularizer}
\end{equation}
which discourages the model from explaining ranking errors through excessive uncertainty alone. We additionally impose a minimum positive reward margin:
\begin{equation}
\mathcal L_{\mathrm{margin}}
=\mathbb E[(m_0-m)_+].
\label{eq:margin_regularizer}
\end{equation}
Here, $(x)_+=\max(x,0)$ and $m_0>0$. The margin term penalizes preferred--rejected pairs whose predicted scores are insufficiently separated. A single pair constrains only the variance sum $S$, so candidate-level variance is not independently identified by one label. Candidate-level structure can arise only from the state-conditioned variance head shared across many training comparisons. Accordingly, downstream weighting normalizes the predicted reliability across samples evaluated at the same transition.
\paragraph{Training objective.}
Pairwise ranking is invariant to adding the same constant to both scores. We use a small quadratic anchor to remove this free offset and stabilize the overall score scale:
\begin{equation}
\mathcal L_{\mathrm{anchor}}
=\tfrac12\mathbb E[(\mu_\theta^+)^2+(\mu_\theta^-)^2].
\label{eq:score_anchor}
\end{equation}
The complete reward-model objective is
\begin{align}
\mathcal L_{\mathrm{RM}}
={}&\mathcal L_{\mathrm{pref}}
+\lambda_{\mathrm{var}}\mathcal L_{\mathrm{var}}
+\lambda_{\mathrm{margin}}\mathcal L_{\mathrm{margin}}
\nonumber\\
&+\lambda_{\mathrm{anchor}}\mathcal L_{\mathrm{anchor}}.
\label{eq:reward_model_objective}
\end{align}
Loss weights and the component-ablation protocol are given in the supplementary material.
\paragraph{Unified model architecture.}
Across image and video modalities, \textsc{SURE-LRM} reuses the DiT backbone of the corresponding generator as its latent feature encoder, providing a unified architecture that directly processes noisy latents, prompts, and timesteps without VAE decoding. For images, learned queries aggregate multi-layer visual--text features; for videos, a query-attention pooler summarizes intermediate spatiotemporal features from the Wan DiT. A shared two-output projection maps the aggregated representation to the mean reward $\mu$ and residual-variance parameter $a$. Backbone-specific feature layers, query counts, and projection dimensions are provided in the supplementary material.

\subsection{\textsc{SURE-REFL}: Uncertainty-Guided Dense Reward Feedback}
\label{sec:uncertainty_guided_posttraining}

After training \textsc{SURE-LRM}, we freeze its parameters. Its mean supplies the optimization signal, while its detached uncertainty determines how strongly each local signal contributes.

\paragraph{Transition-local feedback and reliability.}
Let the sampler contain $T$ denoising transitions, and let $\mathcal J\subseteq\{0,\ldots,T-1\}$ be the queried subset. We write $\operatorname{sg}(\cdot)$ for stop-gradient. At transition $j$, we detach the incoming state and use the trainable generator to predict the current velocity:
\begin{align}
\overline{\mathbf z}_j&=\operatorname{sg}(\mathbf z_j),
\nonumber\\
\mathbf v_j^\phi&=g_\phi(\overline{\mathbf z}_j,c,t_j),
\label{eq:local_velocity_prediction}
\end{align}
where $\mathbf v_j^\phi$ is the velocity predicted by generator $g_\phi$ with trainable parameters $\phi$. The native scheduler or solver then advances the detached state by one denoising transition:
\begin{equation}
\mathbf z_{j+1}=\mathcal S_j(\overline{\mathbf z}_j,\mathbf v_j^\phi).
\label{eq:transition_local_update}
\end{equation}
For minibatch sample $b$, the frozen reward model evaluates this post-transition state and returns its mean reward and unconstrained residual-variance parameter:
\begin{equation}
f_\theta(\mathbf z_{j+1,b},c_b,t_{j+1})
=(\mu_{j,b},a_{j,b}).
\label{eq:posttraining_reward_outputs}
\end{equation}
Its sample-specific residual variance is
\begin{equation}
v_{\eta,\mathrm{res},j,b}
=\operatorname{softplus}(a_{j,b})+\epsilon_v.
\label{eq:posttraining_residual_variance}
\end{equation}
Adding the noise-dependent prior gives the total variance
\begin{equation}
V_{j,b}=v_0(\sigma_{j+1})+v_{\eta,\mathrm{res},j,b}.
\label{eq:posttraining_variance}
\end{equation}
We convert this variance into a detached inverse-standard-deviation reliability:
\begin{equation}
\rho_{j,b}=\operatorname{sg}(V_{j,b}^{-1/2}).
\label{eq:posttraining_reliability}
\end{equation}
Lower predicted variance yields a larger reliability weight. We detach $\mathbf z_{j+1}$ before the next sampling step, confining each reward gradient to its current transition and avoiding retention of the full denoising graph. Detaching $\rho_{j,b}$ leaves the mean reward as the only gradient path.

\paragraph{Unified image--video post-training.}
For both image and video generators, we normalize reliability across samples evaluated at the same transition. Let $\mathcal B$ denote this matched-transition batch and let $b$ index its samples. The mean reliability is
\begin{equation}
\bar\rho_j=
\frac{1}{|\mathcal B|}
\sum_{b\in\mathcal B}\rho_{j,b}
\label{eq:mean_transition_reliability}
\end{equation}
and define the bounded reliability weight
\begin{equation}
\beta_{j,b}=\beta_{\mathrm{base}}+\lambda_p\min\!\left(
\left[\rho_{j,b}/\bar\rho_j-1\right]_+,B_p
\right).
\label{eq:refl_reliability_multiplier}
\end{equation}
Here, $\beta_{\mathrm{base}}\geq0$ controls the minimum reward weight, $\lambda_p\geq0$ sets the strength of reliability modulation, and $B_p>0$ caps its increase. Setting $\beta_{\mathrm{base}}=0$ suppresses reward feedback at or below the matched-transition average, whereas a positive value preserves baseline supervision for every sample. The cap prevents a small number of high-reliability states from dominating the update.

Given transition weights $w_j\geq0$ with $\sum_{j\in\mathcal J}w_j=1$ and reward scale $\eta_R>0$, the shared uncertainty-weighted reward objective is
\begin{equation}
\mathcal L_{\mathrm{REFL}}
=-\eta_R\sum_{j\in\mathcal J}w_j
\mathbb E_b[\beta_{j,b}\mu_{j,b}].
\label{eq:refl_posttraining_objective}
\end{equation}
The same normalization and objective are used for image and video post-training. The modality changes the generator, latent dimensionality, scheduler, and queried transitions, but not the uncertainty-weighting rule.
\section{Experiments}
\label{sec:experiments}

\subsection{Experimental Setup}
\label{sec:experimental_setup}
\begin{table}[t]
\centering
\begingroup
\small
\setlength{\tabcolsep}{1.7pt}
\renewcommand{\arraystretch}{0.98}
\resizebox{\columnwidth}{!}{%
\begin{tabular}{@{}llccccc@{}}
\toprule
\multirow{2}{*}{\textbf{Model}}
& \multirow{2}{*}{\textbf{Backbone}}
& \multicolumn{5}{c}{\textbf{Pairwise Accuracy (\%)}} \\
\cmidrule(lr){3-7}
& & \shortstack{\textbf{Image}\\\textbf{Reward}}
  & \shortstack{\textbf{HPD}\\\textbf{v2}}
  & \shortstack{\textbf{HPD}\\\textbf{v3}}
  & \shortstack{\textbf{GenAI}\\\textbf{Bench}}
  & \textbf{Avg} \\
\midrule
\multicolumn{7}{@{}l}{\textit{CLIP-based Pixel Reward}} \\
\midrule
ImageReward & BLIP ViT-L & 65.15 & 73.95 & 58.74 & 63.41 & 65.31 \\
PickScore & OpenCLIP ViT-H/14 & 62.73 & 79.44 & 65.67 & 69.98 & 69.46 \\
HPSv2 & OpenCLIP ViT-H/14 & 65.62 & 82.58 & 64.69 & 67.62 & 70.13 \\
MPS & CLIP-H (ViT-H/14) & 66.37 & 83.27 & 64.33 & 68.08 & 70.51 \\
\midrule
\multicolumn{7}{@{}l}{\textit{VLM-based Pixel Reward}} \\
\midrule
UnifiedReward & LLaVA-OneVision-7B & 63.82 & 83.10 & 71.96 & 72.38 & 72.82 \\
UnifiedReward-Think & LLaVA-OneVision-7B & 58.54 & 82.70 & 66.07 & 70.91 & 69.56 \\
HPSv3 & Qwen2-VL-7B-Instruct & 67.03 & 85.36 & 76.03 & 70.95 & 74.84 \\
\midrule
\multicolumn{7}{@{}l}{\textit{Diffusion-based Reward}} \\
\midrule
LRM-SD1.5 & Stable Diffusion v1.5 & 59.17 & 72.39 & 54.05 & 60.86 & 61.62 \\
LRM-SDXL & SDXL-base-1.0 & 60.35 & 71.19 & 53.80 & 61.58 & 61.73 \\
DiNa-LRM & SD3.5-Medium & 60.34 & 82.13 & 75.04 & 68.43 & 71.49 \\
\rowcolor{gray!30}
\textsc{SURE-LRM} & SD3.5-Medium & 61.85 & \textbf{83.93} & \textbf{75.11} & \textbf{73.26} & \textbf{73.54} \\
DiNa-LRM & Z-Image-Turbo & 60.13 & 81.75 & 71.58 & 67.21 & 70.17 \\
\rowcolor{gray!30}
\textsc{SURE-LRM} & Z-Image-Turbo & 61.26 & 80.99 & 72.82 & 67.72 & 70.70 \\
DiNa-LRM & FLUX.1-dev & 59.03 & 81.21 & 72.57 & 66.67 & 69.87 \\
\rowcolor{gray!30}
\textsc{SURE-LRM} & FLUX.1-dev & \textbf{62.64} & 82.45 & 74.80 & 68.78 & 72.17 \\
\bottomrule
\end{tabular}%
}
\endgroup
\caption{Pairwise preference accuracy across reward-model families. The best measured result within each family is bold. Results for our method report the mean across three random seeds, with one evaluation per seed. Avg is the unweighted mean of the four benchmarks.}
\label{tab:pairwise_preference_accuracy}
\end{table}
\paragraph{Evaluation protocol.}
We first evaluate \textsc{SURE-LRM} on four image preference benchmarks: ImageReward~\cite{xu2023imagereward}, HPD v2~\cite{hpsv2}, HPD v3~\cite{ma2025hpsv3}, and GenAI-Bench~\cite{jiang2024genaiarena}. We compare it with CLIP-based rewards (ImageReward, PickScore, HPSv2, and MPS), VLM-based rewards (UnifiedReward, UnifiedReward-Think, and HPSv3), and diffusion-based rewards (LRM and DiNa-LRM)~\cite{xu2023imagereward,pickscore,hpsv2,mps,unifiedreward,unifiedrewardthink,ma2025hpsv3,zhang2025lrm,liu2026dina}.

For video preference prediction, we use VisionRewardDB~\cite{xu2024visionreward} and a held-out split of T2V Ranking Human Preferences~\cite{datapointai2026t2vranking}; its source videos were introduced with VideoScore2~\cite{he2025videoscore2}. The baselines cover image reward models and video-specific reward models: HPSv2.1, ImageReward, PickScore, VideoScore, VideoReward, Q-Insight, UnifiedReward-Think, and VisionReward~\cite{hpsv21release,xu2023imagereward,pickscore,he2024videoscore,liu2025videoalign,li2025qinsight,unifiedrewardthink,xu2024visionreward}. Published VisionRewardDB results follow the tie-excluded protocol reported by VQ-Insight~\cite{zhang2026vqinsight}.

For image post-training, we test SD3.5-Medium, Z-Image-Turbo, and FLUX.1-dev. The baselines are Flow-GRPO, LPO, ViPO, and DiNa-LRM~\cite{liu2025flowgrpo,zhang2025lrm,ni2025vipo,liu2026dina}. We report HPSv2.1, ImageReward, and PickScore~\cite{hpsv21release,xu2023imagereward,pickscore}. For video, we post-train Wan2.1-480P-T2V-14B~\cite{wan2025wan} and compare with DanceGRPO, ViPO, VideoDPO, and HY-PRFL on VBench~\cite{xue2025dancegrpo,ni2025vipo,liu2025videodpo,mi2025prfl,huang2023vbench}. The supplementary material provides the human study, computational-cost comparison, and full implementation details.

\paragraph{Implementation.}
All variants of a backbone use matched prompts, sampling budgets, resolutions, and metric implementations. For VisionRewardDB, we exclude 180 human-tie pairs and evaluate the remaining 820 comparisons. Published VisionRewardDB baseline results use the same tie-excluded protocol, while the remaining entries are obtained with our evaluation pipeline. Results for our method in Tables~\ref{tab:pairwise_preference_accuracy}, \ref{tab:uncertainty_reliability}, and \ref{tab:video_reward_accuracy} use three random seeds. Each seed is evaluated once, and we report the mean across the three runs. \textsc{SURE-REFL} queries the frozen \textsc{SURE-LRM} at selected denoising transitions and backpropagates through each local transition. Dataset construction, checkpoint identifiers, optimization hyperparameters, sampling schedules, and compute resources are provided in the supplementary material.
\subsection{Pairwise Reward-Model Accuracy}

Table~\ref{tab:pairwise_preference_accuracy} compares image preference accuracy. \textsc{SURE-LRM} achieves a higher average than the matched DiNa-LRM model on all three backbones. The mean across backbones increases from 70.51 to 72.14, and the largest gain is 2.30 points on FLUX.1-dev. It is therefore the strongest diffusion-native reward model in this comparison. Together with the diagnostics below, these results support our central claim: samples at the same timestep can require different reliability estimates, and this information can be learned from pairwise preferences.
\begin{table}[t]
\centering
\begingroup
\scriptsize
\setlength{\tabcolsep}{1.5pt}
\renewcommand{\arraystretch}{1.00}
\begin{tabular*}{\columnwidth}{@{\extracolsep{\fill}}cccc@{}}
\toprule
\shortstack{\textbf{Pair Acc.}\\\textbf{100\% / 50\%}}
& \shortstack{\textbf{Mean Residual Var.}\\\textbf{Correct / Error}}
& \shortstack{\textbf{Error Var. Increase}\\\textbf{(95\% CI)}}
& \shortstack{\textbf{Error AUROC}\\\textbf{(95\% CI)}} \\
\midrule
79.40\% / 90.10\%
& 0.1724 / 0.2142
& \shortstack{24.26\%\\{[20.44, 28.07]}}
& \shortstack{0.6941\\{[0.6653, 0.7224]}} \\
\midrule
\multicolumn{4}{@{}l}{\textit{Agreement with human confidence}} \\
\midrule
\textbf{Condition}
& \shortstack{\textbf{Pearson}\\$r$}
& \shortstack{\textbf{Spearman}\\$\rho$}
& \shortstack{\textbf{Mean Residual Var.}\\\textbf{Low / High Conf.}} \\
\midrule
$u=0$   & 0.682 & 0.796 & 0.2367 / 0.0670 \\
$u=0.3$ & 0.672 & 0.785 & 0.3022 / 0.0589 \\
\bottomrule
\end{tabular*}
\endgroup
\caption{Uncertainty diagnostics. The upper block evaluates a subset of 2,000 pairs; the 50\% result retains the half with the lowest predicted residual variance. The lower block reports human-confidence agreement and the lowest/highest confidence bins. Complete binwise results and definitions of the $u$ conditions are provided in the supplementary material. Results report the mean across three random seeds, with one evaluation per seed.}
\label{tab:uncertainty_reliability}
\end{table}

\paragraph{Uncertainty reliability.}
At a fixed transition, all samples share the same noise-dependent prior. Their variance differences therefore come from the learned residual term. Table~\ref{tab:uncertainty_reliability} shows that incorrect rankings have 24.26\% higher residual variance on average. Predicted variance identifies ranking errors with an AUROC of 0.6941, and retaining the lowest-variance half raises pairwise accuracy from 79.40\% to 90.10\%. It also agrees with human confidence under both $u$ conditions. These results show that the learned variance contains useful information about relative reward reliability, which motivates its use in \textsc{SURE-REFL}. Because these pairs appeared during training, we treat this evidence as an in-distribution diagnostic rather than held-out calibration.

\begin{table}[t]
\centering
\begingroup
\small
\setlength{\tabcolsep}{1.8pt}
\renewcommand{\arraystretch}{0.98}
\resizebox{\columnwidth}{!}{%
\begin{tabular}{@{}llcc@{}}
\toprule
\multirow{2}{*}{\textbf{Reward Model}}
& \multirow{2}{*}{\textbf{Backbone}}
& \multicolumn{2}{c}{\textbf{Pairwise Accuracy (\%)}} \\
\cmidrule(lr){3-4}
& & \textbf{VisionRewardDB} & \textbf{T2V Ranking} \\
\midrule
HPSv2.1 & OpenCLIP ViT-H/14 & 63.17 & 56.04 \\
ImageReward-v1.0 & BLIP ViT-L & 62.07 & 57.57 \\
PickScore-v1 & OpenCLIP ViT-H/14 & 62.44 & 62.28 \\
\midrule
VideoScore & Mantis-Idefics2-8B & 54.90 & 54.68 \\
VideoReward & Qwen2-VL-2B-Instruct & 59.88 & 58.75 \\
Q-Insight & Qwen2.5-VL-7B-Instruct & 60.37 & 61.24 \\
UnifiedReward-Think & LLaVA-OneVision-7B & 62.56 & 63.84 \\
VisionReward & CogVLM2-Video-LLaMA3-Chat & 72.03 & 71.64 \\
\midrule
\rowcolor{gray!10}
\textsc{SURE-LRM} & Wan2.1-480P-T2V-14B & \textbf{72.31} & \textbf{71.70} \\
\bottomrule
\end{tabular}%
}
\endgroup
\caption{Video preference prediction accuracy. The best result on each benchmark is shown in bold. Results for our method report the mean across three random seeds, with one evaluation per seed.}
\label{tab:video_reward_accuracy}
\end{table}

Table~\ref{tab:video_reward_accuracy} tests whether the same latent reward formulation transfers to video. \textsc{SURE-LRM} obtains the best measured accuracy on both benchmarks: 72.31\% on VisionRewardDB and 71.70\% on T2V Ranking. The gains over VisionReward are modest, but they show that a reward model operating directly on video latents can match or exceed strong decoded-video evaluators. This result supports the use of one sample-adaptive reward formulation across image and video backbones.

\begin{table}[!t]
\centering
\begingroup
\footnotesize
\setlength{\tabcolsep}{2.0pt}
\renewcommand{\arraystretch}{0.95}
\begin{tabularx}{\columnwidth}{
  @{}
  l
  *{3}{>{\centering\arraybackslash}X}
  @{}
}
\toprule
\multirow{2}{*}{\textbf{Method}}
& \multicolumn{3}{c}{\textbf{Automatic Reward Score}} \\
\cmidrule(lr){2-4}
& \shortstack{\textbf{HPS}\\\textbf{v2.1}$\uparrow$}
& \shortstack{\textbf{Image}\\\textbf{Reward}$\uparrow$}
& \shortstack{\textbf{Pick}\\\textbf{Score}$\uparrow$} \\
\midrule

\multicolumn{4}{@{}l}{\textit{SD3.5-Medium}} \\
Base
& 0.2890 & 1.068 & 22.40 \\
Flow-GRPO
& 0.3024 & 1.112 & 22.46 \\
LPO
& 0.2944 & 1.045 & 22.34 \\
ViPO
& \underline{0.3223}
& \underline{1.318}
& \underline{22.71} \\
DiNa-LRM
& 0.3146 & 1.200 & 22.61 \\
\rowcolor{gray!10}
\textsc{SURE-REFL}
& \textbf{0.3375}
& \textbf{1.370}
& \textbf{22.73} \\

\midrule
\multicolumn{4}{@{}l}{\textit{Z-Image-Turbo}} \\
Base
& 0.2817 & 0.8644 & 22.48 \\
Flow-GRPO
& 0.2886 & 0.1022 & 22.28 \\
LPO
& 0.2848 & 0.8783 & 22.03 \\
ViPO
& \underline{0.3018}
& \underline{1.115}
& \underline{22.72} \\
DiNa-LRM
& 0.2992 & 0.9187 & 22.54 \\
\rowcolor{gray!10}
\textsc{SURE-REFL}
& \textbf{0.3294}
& \textbf{1.265}
& \textbf{22.77} \\

\midrule
\multicolumn{4}{@{}l}{\textit{FLUX.1-dev}} \\
Base
& 0.2920 & 0.8566 & 22.36 \\
Flow-GRPO
& \underline{0.3127} & 0.9882 & 22.18 \\
LPO
& 0.2848 & 0.8783 & 22.03 \\
ViPO
& 0.3025 & 0.9513 & \textbf{22.51} \\
DiNa-LRM
& 0.3102 & \underline{1.093} & 22.47 \\
\rowcolor{gray!10}
\textsc{SURE-REFL}
& \textbf{0.3351}
& \textbf{1.126}
& \underline{22.49} \\

\bottomrule
\end{tabularx}
\endgroup
\caption{Quantitative evaluation over SOTA baselines on various matrics. SURE-REFL achieved leading image post-train performance. The best values are shown in bold, and the second-best are underlined.}
\label{tab:post_training_comparison}
\end{table}

\begin{table*}[!t]
\centering
\begingroup
\scriptsize
\setlength{\tabcolsep}{0.4pt}
\renewcommand{\arraystretch}{1.00}
\begin{tabularx}{\textwidth}{@{}
>{\raggedright\arraybackslash}p{16mm}
*{15}{>{\centering\arraybackslash}X}
@{}}
\toprule
\multirow{2}{*}{\textbf{Method}}
& \multicolumn{6}{c}{\textbf{Selected Quality Dimensions}}
& \multicolumn{6}{c}{\textbf{Selected Semantic Dimensions}}
& \multicolumn{3}{c}{\textbf{Official Aggregates}} \\
\cmidrule(lr){2-7}\cmidrule(lr){8-13}\cmidrule(lr){14-16}
& \textbf{Subj.} & \textbf{Back.} & \textbf{Motion} & \textbf{Dynamic} & \textbf{Aesth.} & \textbf{Imaging}
& \textbf{Object} & \textbf{Multi.} & \textbf{Action} & \textbf{Spatial} & \textbf{Scene} & \textbf{Temp.}
& \textbf{Quality} & \textbf{Semantic} & \textbf{Total} \\
\midrule
Base
& 0.9647 & 0.9731 & 0.9848 & 0.5278 & 0.6141 & 0.6790
& 0.8180 & 0.6996 & 0.7700 & 0.7294 & 0.3249 & 0.2412
& 0.8336 & 0.7120 & 0.8092 \\
DanceGRPO
& \textbf{0.9769} & \textbf{0.9779} & 0.9879 & 0.4583 & 0.6268 & 0.6931
& 0.8156 & 0.6318 & 0.8100 & 0.7118 & 0.2914 & 0.2414
& 0.8378 & 0.6969 & 0.8096 \\
ViPO
& 0.9621 & 0.9684 & 0.9823 & \textbf{0.6527} & 0.6114 & 0.6827
& 0.8338 & 0.6958 & 0.7900 & 0.7307 & 0.2725 & 0.2407
& 0.8390 & 0.7101 & 0.8132 \\
Video-DPO
& 0.9527 & 0.9728 & 0.9861 & 0.6228 & 0.6533 & 0.6449
& 0.8922 & 0.7089 & 0.8108 & 0.7421 & 0.3474 & 0.2401
& 0.8445 & 0.7187 & 0.8248 \\
HY-PRFL
& 0.9721 & 0.9745 & 0.9834 & 0.6068 & 0.6408 & 0.6922
& 0.8453 & 0.7224 & 0.8004 & 0.7536 & 0.3188 & 0.2442
& 0.8412 & 0.7256 & 0.8248 \\
\rowcolor{gray!10}
\textsc{SURE-REFL}
& 0.9665 & 0.9743 & \textbf{0.9882} & 0.6388 & \textbf{0.6720} & \textbf{0.7026}
& \textbf{0.9280} & \textbf{0.8041} & \textbf{0.8200} & \textbf{0.8621} & \textbf{0.3968} & \textbf{0.2462}
& \textbf{0.8548} & \textbf{0.7596} & \textbf{0.8357} \\
\bottomrule
\end{tabularx}
\endgroup
\caption{Quantitative evaluation over SOTA baselines on VBench. SURE-REFL achieved leading video post-train performance. }
\label{tab:vbench_method_comparison}
\end{table*}

\begin{figure*}[!t]
    \centering
    \includegraphics[width=\textwidth]{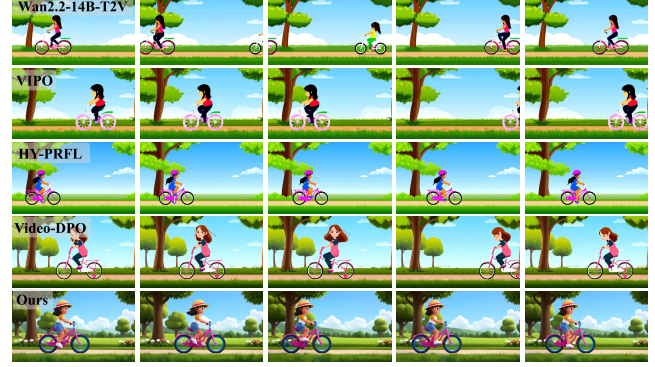}
    \caption{Qualitative comparison on video results of \textsc{SURE-REFL}.}
    \label{fig:video_qualitative_cases}
\end{figure*}

\subsection{Image-Model Post-Training}

\begin{figure}[!t]
    \centering
    \includegraphics[width=\columnwidth]{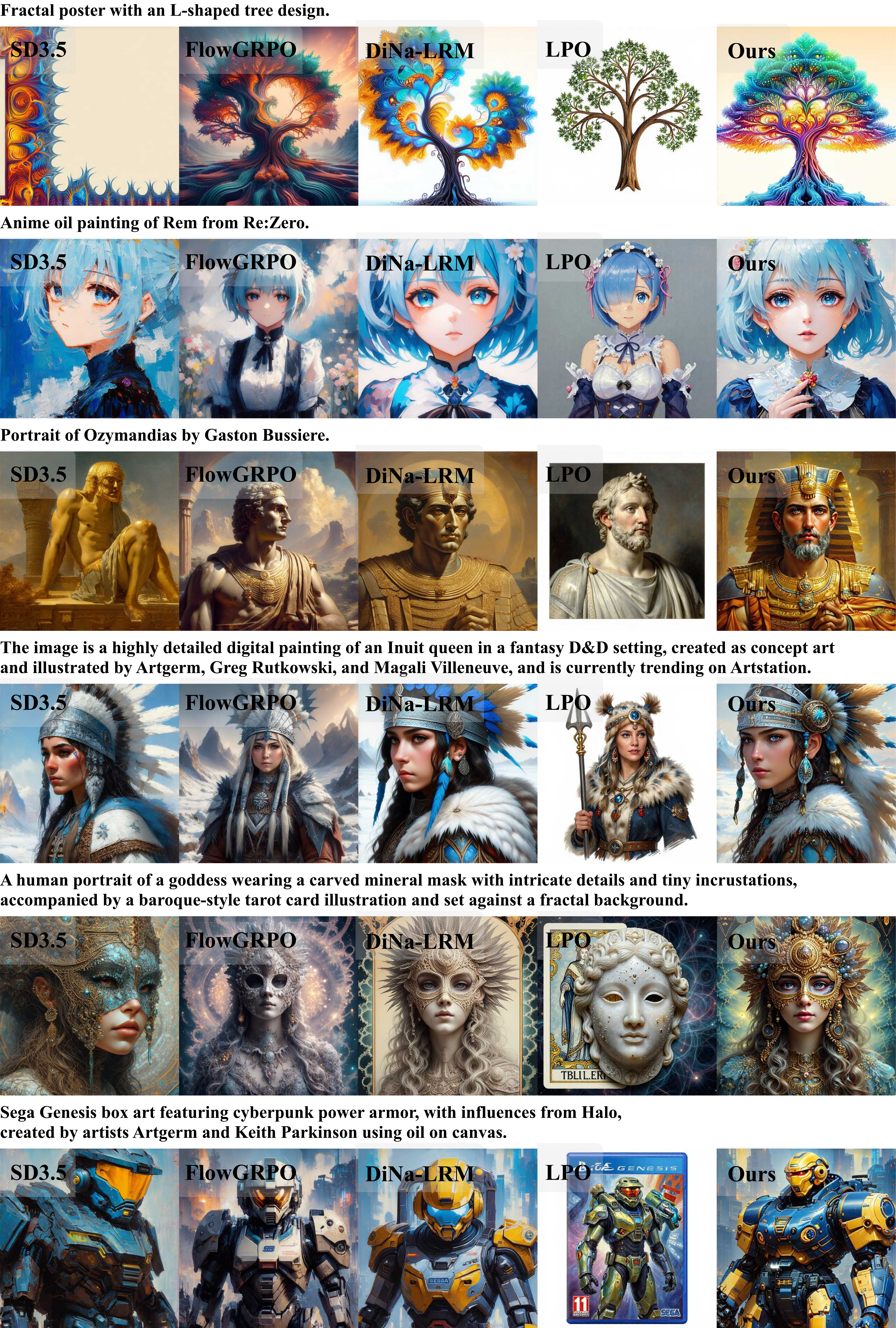}
    \caption{Qualitative comparison on image results of \textsc{SURE-REFL}.}
    \label{fig:image_qualitative_cases}
\end{figure}

Table~\ref{tab:post_training_comparison} evaluates image post-training on three different generators. \textsc{SURE-REFL} ranks first in eight of the nine backbone--metric comparisons and improves HPSv2.1 on every backbone. The gains hold across backbones rather than depending on a single generator or metric. \textsc{SURE-REFL} compares reliability only among samples at the same transition, so the shared noise level cannot dominate the weights. It can then use dense latent feedback while reducing the influence of ambiguous reward predictions. Matched-prompt examples are shown in Figure~\ref{fig:image_qualitative_cases}.It demonstrates that our method yields stable and significant performance gains for image downstream training.

\begin{figure}[t]
    \centering
    \includegraphics[width=0.94\columnwidth]{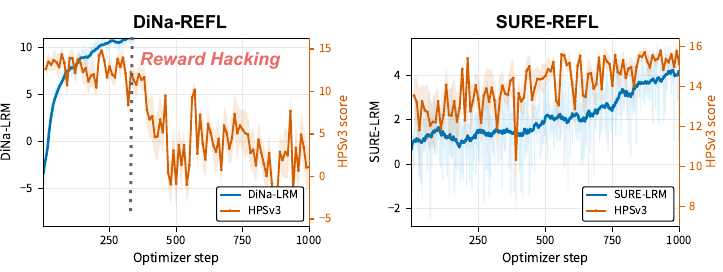}
    \caption{Optimization dynamics of DiNa-LRM-REFL (left) and \textsc{SURE-REFL} (right). The former shows reward hacking as its optimized reward rises while HPSv3 falls. \textsc{SURE-REFL} remains stable for longer and reaches better scores.}
    \label{fig:reward_hacking_dynamics}
\end{figure}

\paragraph{Optimization stability.}
Figure~\ref{fig:reward_hacking_dynamics} compares the optimized reward with an independent HPSv3 score. For DiNa-LRM-REFL, the optimized reward keeps rising after HPSv3 begins to fall, indicating reward hacking. With \textsc{SURE-REFL}, the two signals remain aligned for longer and reach better final values. This result supports our claim that uncertainty-guided weighting reduces harmful updates from ambiguous latent feedback.

\subsection{Video Post-Training on VBench}

Table~\ref{tab:vbench_method_comparison} evaluates video post-training. \textsc{SURE-REFL} achieves the highest quality, semantic, and total VBench scores. Its total score increases from 0.8092 to 0.8357 and is 0.0109 above the strongest baseline. The image and video implementations use the same matched-transition weighting rule. The result therefore supports our claim that uncertainty-guided dense feedback transfers across modalities without changing the core objective. Figure~\ref{fig:video_qualitative_cases} provides qualitative video comparisons.  
It demonstrates that SURE‑LRM can consistently improve image quality and motion smoothness, while also enhancing text alignment, yielding stable and significant performance gains for video post‑training.

\begin{table}[t]
\centering
\begingroup
\scriptsize
\setlength{\tabcolsep}{1.5pt}
\renewcommand{\arraystretch}{0.96}
\begin{tabular}{@{}lccccc@{}}
\toprule
\multirow{2}{*}{\textbf{Loss Configuration}}
& \multicolumn{5}{c}{\textbf{Pairwise Accuracy (\%)}} \\
\cmidrule(lr){2-6}
& \shortstack{\textbf{Image}\\\textbf{Reward}}
& \shortstack{\textbf{HPD}\\\textbf{v2}}
& \shortstack{\textbf{HPD}\\\textbf{v3}}
& \shortstack{\textbf{GenAI}\\\textbf{Bench}}
& \textbf{Avg} \\
\midrule
$\mathcal L_{\mathrm{pref}}$ only & 50.14 & 51.28 & 54.14 & 52.43 & 52.00 \\
w/o $\mathcal L_{\mathrm{var}}$ & 56.21 & 79.14 & 70.25 & 69.27 & 68.72 \\
w/o $\mathcal L_{\mathrm{margin}}$ & 58.14 & 81.26 & 73.24 & 72.98 & 71.41 \\
w/o $\mathcal L_{\mathrm{anchor}}$ & 60.14 & 80.24 & 73.87 & 72.06 & 71.58 \\
\rowcolor{gray!10}
\textbf{Full objective} & 61.85 & 83.93 & 75.11 & 73.26 & 73.54 \\
\bottomrule
\end{tabular}%
\endgroup
\caption{Ablation of reward-model training losses on SD3.5-Medium. }
\label{tab:sure_loss_ablation}
\end{table}

\paragraph{Loss ablation.}
Table~\ref{tab:sure_loss_ablation} shows that all parts of the reward-model objective are useful. The preference loss alone remains close to chance. Removing $\mathcal L_{\mathrm{var}}$ causes the largest leave-one-out drop, from 73.54 to 68.72, confirming the need to control variance inflation. The margin loss keeps preferred and rejected scores separated, while the anchor fixes their otherwise free common offset. The full objective performs best on all four benchmarks. Additional ablations are provided in the supplementary material.

\section{Conclusion}
\label{sec:conclusion}

\textsc{SURE} unifies uncertainty-aware latent reward learning and dense visual-model alignment. \textsc{SURE-LRM} learns candidate-dependent reward variance directly from pairwise preferences, and \textsc{SURE-REFL} converts this uncertainty into bounded, detached weights for local image and video post-training. \textsc{SURE-LRM} improves matched-backbone pairwise accuracy over DiNa-LRM, while \textsc{SURE-REFL} yields consistent image-generation gains and the highest VBench aggregate scores among the compared methods. Future work should examine uncertainty calibration under distribution shift and extend the evaluation to additional video domains. Overall, the results support treating state-specific reliability and local trajectory feedback as first-class properties of latent rewards used for visual alignment.

\bibliography{aaai2027}

\end{document}